\documentclass[conference]{IEEEtran}

\IEEEoverridecommandlockouts
\usepackage{cite}
\usepackage{amsmath,amssymb,amsfonts}
\usepackage{amsthm}
\usepackage{algorithmic}
\usepackage{graphicx}
\usepackage{textcomp}
\usepackage{xcolor}
\usepackage[]{footmisc}
\usepackage{bbm}

\usepackage{cuted}
\usepackage{stfloats}
\usepackage{mathtools,amssymb,lipsum, nccmath}
\newtheorem{theorem}{Theorem}

\usepackage{algorithm,algorithmic}
\usepackage{color,soul}
\usepackage{graphicx}
\usepackage{subfig}

\def\BibTeX{{\rm B\kern-.05em{\sc i\kern-.025em b}\kern-.08em
    T\kern-.1667em\lower.7ex\hbox{E}\kern-.125emX}}
\begin{document}

\title{Analog-digital Scheduling for Federated Learning: A Communication-Efficient Approach 
\\
\thanks{M. Faraz and N. Michelusi are with the School of Electrical, Computer and Energy
Engineering, Arizona State University. email: \{mulabrar,
nicolo.michelusi\}@asu.edu.
\\
This research has been funded in part by NSF under grant CNS-$2129615$.}
}

\author{\IEEEauthorblockN{Muhammad Faraz Ul Abrar}
\and
\IEEEauthorblockN{Nicolò Michelusi}
}

\maketitle
\setulcolor{red}
\setul{red}{2pt}
\setstcolor{red}

\begin{abstract}
Over-the-air (OTA) computation has recently emerged as a communication-efficient Federated Learning (FL) paradigm to train machine learning models over wireless networks. However, its performance is limited by the device with the worst SNR, resulting in fast yet noisy updates. On the other hand, allocating orthogonal resource blocks (RB) to individual devices via digital channels mitigates the noise problem, at the cost of increased communication latency. In this paper, we address this discrepancy and present ADFL, a novel Analog-Digital FL scheme: in each round, the parameter server (PS) schedules each device to either upload its gradient via the analog OTA scheme or transmit its quantized gradient over an orthogonal RB using the ``digital" scheme. Focusing on a single FL round, we cast the optimal scheduling problem as the minimization of the mean squared error (MSE) on the estimated global gradient at the PS, subject to a delay constraint, yielding the optimal device scheduling configuration and quantization bits for the digital devices. Our simulation results show that ADFL, by scheduling most of the devices in OTA scheme while also occasionally employing the digital scheme for a few devices, consistently outperforms OTA-only and digital-only schemes, in both i.i.d. and non-i.i.d. settings. \end{abstract}




\section{Introduction}
The enormous increase in the number of Internet-of-Things (IoT) devices generating data traffic has promoted the widespread adoption of distributed learning-based solutions. Federated Learning (FL) has emerged as a promising candidate to address the privacy issues and alleviate the communication burden in distributed learning \cite{DistL_intro}. With FL, $N$ devices collaborate with a central PS by only exchanging local parameter or gradient information \cite{FL_intro,MH_FL,DistL_intro,FL_survey}.
A standard FL setting aims to learn a global model as
\begin{equation}
\boldsymbol{\theta}^* = \arg\min_{\theta \in \mathbb{R}^d} F(\boldsymbol{\theta}) \triangleq \frac{1}{N} \sum_{m \in \mathcal{N}} f_m(\boldsymbol{\theta}), \tag{P}
\label{ML_prob}
\end{equation}
where $f_m(\boldsymbol{\theta})$ is the local function of the $m$-th device. Typically, \eqref{ML_prob} is solved using distributed stochastic gradient descent (SGD) in an iterative fashion: the devices compute local gradients using their local datasets, and upload them to the PS; the PS then updates the global model by averaging the gradients and delivers it to the devices. This process continues over multiple rounds until convergence is achieved \cite{Fedavg}. 

Yet, realizing a real-world FL solution requires communications over unreliable wireless channels. Therefore, in practice, FL updates may be significantly affected by fading channels, low signal-to-noise ratio (SNR), and interference in the wireless medium. 
The work in \cite{Joint_L_Comm} shows that FL convergence in a wireless setting is affected by transmit power, RB allocation, and device selection due to the limited wireless spectrum. Similarly, \cite{Delay_aware_FL} investigates FL accounting for the communication delays, whereas \cite{Lat_cons_FL} studies a latency-constrained FL setting over wireless networks. Several works \cite{FL_fading, Prob_Quant1, Prob_Quant2} have proposed sparsification and quantization schemes to realize faster updates. While these techniques help reduce the communication overhead in uploading high dimensional local parameters or gradients, nevertheless, scheduling a large number of devices over orthogonal RBs in each round is impractical owing to spectrum resource scarcity. \cite{Sched_policies}.



To overcome this challenge and ensure scalable device participation, several works \cite{OTA_FL,FL_fading, BB_FL} have proposed analog over-the-air (OTA) computation for FL-based solutions. 
 OTA computation leverages the waveform superposition property of wireless multiple access channels (MAC), making the per-round communication latency independent of the number of devices. However, a key requirement for unbiased OTA aggregation is the coherent ``alignment" of the aggregating uncoded analog modulated signals, making the receive SNR dominated by the \textit{weakest} device \cite{BB_FL}, and thereby, FL updates more susceptible to noise. Consequently, there exists a fundamental trade-off between accuracy and latency in FL realized over wireless channels \cite{Sched_latency_FL,Sched_latency_quant_FL}.

Recognizing this trade-off, we propose ADFL,\footnote{We would like to emphasize that the proposed scheme strikes a balance between MSE and latency, and therefore, can be applied to a more general class of problems, e.g., distributed processing and distributed inference.} which leverages both analog OTA aggregation to support a large number of devices, and digital transmissions over orthogonal RBs to achieve reliable FL updates.
Since the device scheduling directly impacts the latency-noise trade-off \cite{Sched_latency_FL}, several works have proposed scheduling schemes for FL \cite{Sched_policies, BB_FL, Upd_aware_sched, FL_fading}.
Nevertheless, all of these works either consider digital-only, or OTA-only, but not both transmission schemes. In contrast, with ADFL, we minimize this trade-off by optimizing the transmission scheme selection for each device (OTA or digital). Specifically, in each ADFL round, the PS decides the transmission scheme of each device to minimize the MSE under a given round delay requirement. 
Furthermore, for the digitally scheduled devices, we employ dithered quantization and achieve communication efficiency by optimizing the number of bits (i.e., quantization levels). We show that, while the bit allocation problem can be solved efficiently using convex optimization techniques, the device scheduling is a combinatorial optimization problem. By leveraging its structure, we show that it can be converted into a linear search program.
Finally, we demonstrate numerically that, with a suitable learning stepsize, ADFL consistently achieves better test accuracy than existing schemes in different settings. 

The rest of the paper is organized as follows. In Sec. II, we describe ADFL, followed by the device scheduling and digital bit allocation optimization in Sec. III. In Sec. IV, we present numerical results. Finally, Section V concludes the paper.

\textit{Notation}: We denote the set of real and complex numbers by $\mathbb{R}$ and $\mathbb{C}$, respectively. A vector is represented using bold-face lowercase letters. A zero mean circularly symmetric complex Gaussian distributed random variable with variance $\sigma^2$ is denoted as $ \mathcal{CN}$(0, $\sigma^2$). $\Vert \boldsymbol{x} \Vert_\infty $ represents the $\ell$-infinity norm computed as: $ \max\limits_i \vert x_i \vert$. $\lfloor x \rfloor$ denotes the floor operation. The indicator function is represented using $\chi(\cdot)$. 
\section{System Model and analog-digital FL}
\begin{figure}[t!]
	\centering
	\includegraphics[width=0.45\textwidth]{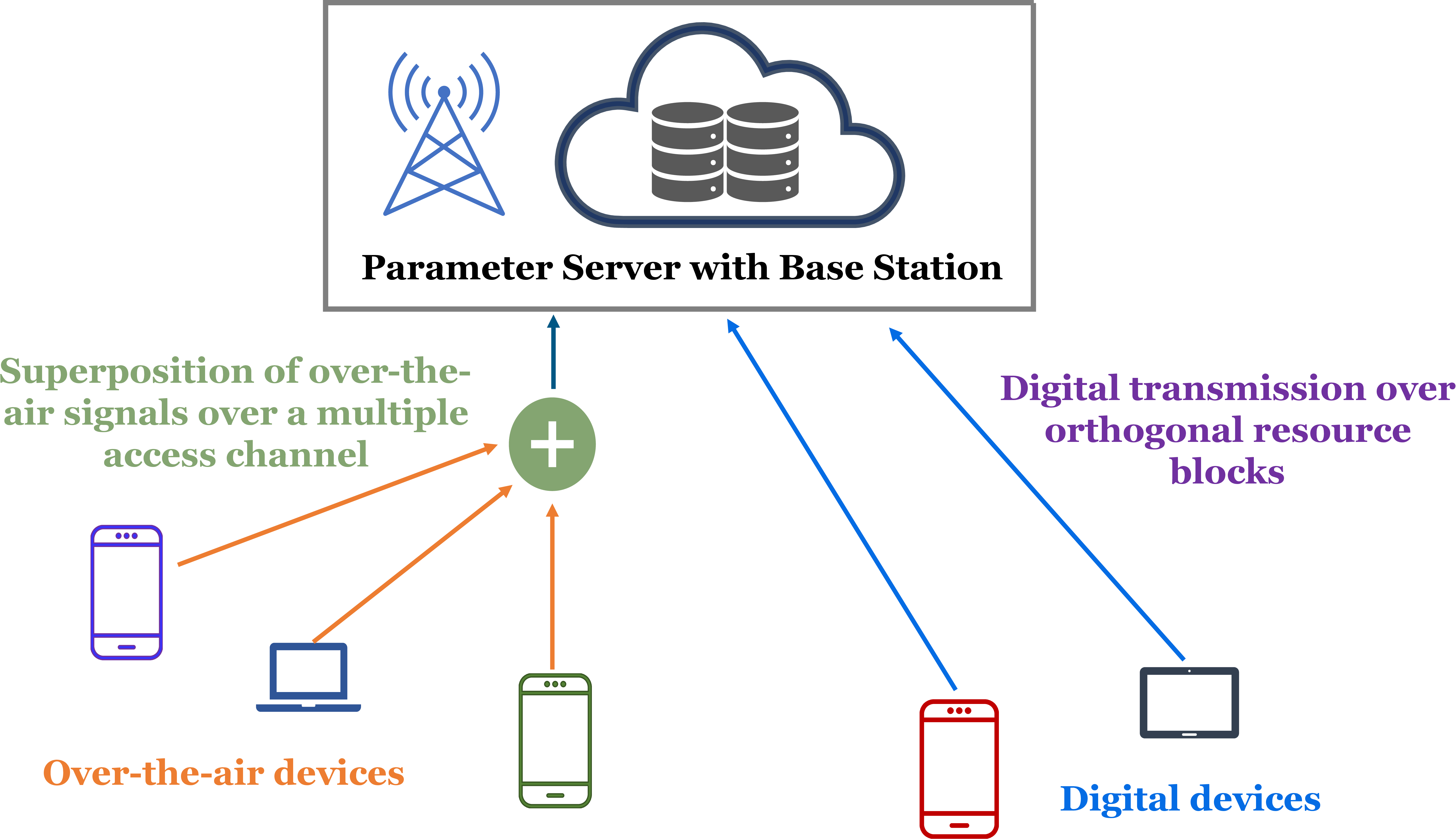}
	\caption{Illustration of ADFL system model}
\label{System_model}
\end{figure}

We consider the setting depicted in Fig. \ref{System_model}, where a distributed wireless network of $N$ devices collaborates with a PS in a FL setting to learn a global model parameter. The $m$-th device has its local dataset $\mathcal{D}_m = \{(\boldsymbol{x}^{(1)}_m, y^{(1)}_m), (\boldsymbol{x}^{(2)}_m,y^{(2)}_m), \cdots \}$, where $\boldsymbol{x}^{(i)}_m$ and $y^{(i)}_m$ are the feature vector and class label, respectively, of the $i$-th local data point. Associated with this dataset, we define the local function $f_m(\boldsymbol{\theta}) = \frac{1}{\vert D_m \vert} \,\sum_{(\boldsymbol{x}_m^{(i)},y_m^{(i)}) \in \mathcal{D}_m} \phi(\boldsymbol{\theta},(\boldsymbol{x}_m^{(i)},y_m^{(i)}))$, only known to device $m$, where $\phi(\boldsymbol{\theta},\cdot)$ is the loss function and $\boldsymbol{\theta} \in \mathbb{R}^d$ is the learning parameter.
We assume that mini-batch parallel SGD is used to solve \eqref{ML_prob} over multiple FL rounds.
Specifically, in the $t$-th FL round, the PS broadcasts the model parameter $\boldsymbol{\theta}_t$
to each device over an error-free downlink channel (as also assumed in \cite{FL_fading, Upd_aware_sched,Sched_policies}); upon receiving $\boldsymbol{\theta}_t$, each device randomly samples a mini-batch $\mathcal{B}_m{\subseteq}\mathcal{D}_m$ of local data points to compute the local gradient $\boldsymbol{g}_{m,t}{=}\frac{1}{\vert\mathcal{B}_m\vert} \sum_{(\boldsymbol{x}_m^{(i)},y_m^{(i)}) \in \mathcal{B}_m}\nabla\phi(\boldsymbol{\theta}_t,(\boldsymbol{x}_m^{(i)},y_m^{(i)}))$, where $\mathbb{E}[\boldsymbol{g}_{m,t}]{=}\nabla f_m(\boldsymbol{\theta}_t)$
and transmits it to the PS. The PS, upon receiving the local gradients from each device, updates the global parameter $\boldsymbol{\theta}_{t+1}$ as
\begin{align}
	    \boldsymbol{\theta}_{t+1} = \boldsymbol{\theta}_{t} - \eta_t \boldsymbol{g}_t , \quad \boldsymbol{g}_t = \frac{1}{N}\sum_{m \in \mathcal{N}} \boldsymbol{g}_{m,t}, 
      \label{GD}
\end{align}   
where $\eta_t$ represents the learning stepsize. Note that in the above, the local gradients are perfectly \textit{aggregated}, since the gradients are transmitted without error. In practice, however, these gradients are transmitted over wireless channels, resulting in noisy estimates of the global parameter. For the rest of the paper, we consider a particular FL round and, therefore, omit the subscript $t$ for notational convenience.

\subsection{Transmission over the wireless channel and ADFL}
In a wireless networked FL setting, each device uploads its locally computed gradient $\boldsymbol{g}_{m}$ over a wireless channel, which we model as a Rayleigh flat fading channel $h_{m} \sim \mathcal{C N} \left(0,\Lambda_m\right)$, where $\Lambda_m$ represents the average path loss. We also assume that
 $h_m$ does not change during one FL round, and
the large-scale fading ($\Lambda_m$) does not vary over the FL running time. However, unlike most prior works \cite{FL_fading,Blind_FL,FL_S_CSI,BB_FL} that assume the average path losses to be the same across devices ($\Lambda_m = \Lambda,\forall m$), here we assume
a more practical setting in which they may differ.


In the $t$-th ADFL round, the PS schedules each device for the uplink transmission using either analog OTA aggregation or digital channels. For brevity, we term the devices scheduled to use OTA aggregation as ``OTA devices," while those transmitting over digital channels as ``digital devices." In ADFL, the PS makes the scheduling decision using a scheduling metric (SM) that we define later (c.f. \eqref{SM_def}), which each device uploads in an error-free way to the PS before transmitting $\boldsymbol{g}_{m}$. The uploading of the SM by the devices and the broadcast of the scheduling decision by the PS can be realized with an initial signaling phase at the start of each round without significant overhead.

We now describe the uplink signals of the devices in ADFL. Each OTA device transmits uncoded \textit{pre-scaled} gradient over a fading MAC to the PS in a time-synchronous fashion, whereas each digital device uses capacity-achieving error-correcting codes to transmit its local gradient over an orthogonal RB. 
Let $\mathcal{K}$ be the set of digital devices and $\mathbf{x}_{m}$ be the signal transmitted by OTA devices $m\in \mathcal{N}\setminus \mathcal{K}$
, then the signal model for OTA devices can be expressed as
\begin{align}
\mathbf{y} = \sum_{m \in \mathcal{N}\setminus \mathcal{K}} h_{m} \cdot \mathbf{x}_{m} \;+    
\;\mathbf{z},
\label{Signal_model}
\end{align}

where 
$\mathbf{z} \sim \mathcal{C N}(\mathbf{0},N_0 \mathbf{I})$
is the additive white noise at the PS with variance $N_0$.
We let each OTA device perform
channel inversion for coherent combining of local gradients at the PS.\footnote{The individual channel gain $h_{m}$ can be estimated at the device by broadcasting a common pilot signal by the PS assuming a reciprocal nature of the channel.} For the $m$-th OTA device, the transmission signal $\mathbf{x}_{m} \in \mathbb{C}^d$ is therefore given as $\mathbf{x}_{m} =  \frac{\gamma} { {h}_{m}}\boldsymbol{g}_{m}$. Here, $\gamma$ denotes the OTA aggregation \textit{pre-scaler}, which is common to all the devices scheduled for OTA transmission in a round. Here, instead of using a \textit{truncated channel inversion} approach as done in \cite{BB_FL}, we instead leverage the additional degrees of freedom brought on by digital channels to let each OTA device perform channel inversion. This approach has the advantage of ensuring uniform participation of each device and therefore yields an unbiased estimate of the global gradient at the PS. For the uplink transmitting signal $\mathbf{x}_{m}$ of OTA devices, we consider the sample energy constraint, i.e., $\vert x^i_{m}\vert^2 \leq E_s \, , \forall i \in \{1, \cdots , d\},$ where $x^i_{m}$ is $i$-th sample of $\mathbf{x}_{m}$.
Hence, to meet the energy constraint at each device, $\gamma$ is chosen as
\begin{align}
\gamma = \min_{m\in\mathcal{N}\setminus \mathcal{K}} \left\{ \sqrt{E_s}\frac{ \left\vert h_{m} \right\vert} { \left\Vert \boldsymbol{{g}}_{m}\right\Vert_\infty } \right\}.
\label{gamma_def}
\end{align}
Upon receiving $\mathbf y$ as in \eqref{Signal_model},
the PS computes the noisy sum of local gradients uploaded by OTA devices as $\boldsymbol{v}  = \frac{\mathbf{y}}{\gamma}$. Since all OTA devices upload simultaneously, the time taken for the OTA transmission of the $d$-dimensional gradient vectors is $\tau_\text{OTA} = \frac{d}{B}$, where $B$ is the bandwidth of the system.
 
Inspired by \cite{Prob_Quant1,Prob_Quant2}, ADFL supports quantization for digital transmissions. To this end, each digital device first normalizes the gradient as $\Tilde{\boldsymbol{g}}_{m} =\frac{\boldsymbol{g}_{m} }{ \Vert \boldsymbol{g}_{m}\Vert_\infty }$, and uses 
a stochastic quantizer
to quantize each entry using $r_{m}$ bits. Let 
$\mathcal{Q}= \{l^1_m, \cdots, \l^Q_m\}$ 
denote the set of quantization levels associated with the quantizer with 
$Q = 2^{r_m}$ levels uniformly spaced in the interval $[-1,1]$, where $l_m^i \leq l_m^j$ for $i \leq j$. Then, $\Tilde{g}_{m,i} \in [l^j_{m}, l^{j+1}_{m} ]$ is quantized as
  \begin{equation}
 \Tilde{g}^q_{m,i} =\begin{cases} 
l^j_{m} & \text{with probability}\quad 1-p_{m},\\
 l^{j+1}_{m} & \text{otherwise,}\\
 \end{cases}
\end{equation}
where $0 \leq p_{m} \leq 1$ is chosen to make the quantizer unbiased, i.e. $\mathbb{E}[\Tilde{g}^q_{m,i}\mid \Tilde{g}_{m,i}] = \Tilde{{g}}_{m,i} \; \forall \,i \in \{1,\cdots,d\}$. The total payload of $L_m = 64 + d r_{m}$ bits is then transmitted over the allocated RB using capacity-achieving codes, where 64 bits are used for the local gradient norm $\Vert \boldsymbol{g}_{m}\Vert_\infty$. Hence, the transmission rate for a digital device is given as $B \log_2\left(1 + \text{SNR}_{m}\right)$, resulting in a transmission time of duration $\frac{L_m}{B\log_2\left(1 + \text{SNR}_{m}\right)}$, where $\text{SNR}_{m} = \frac{E_s \vert h_{m}\vert^2}{N_0}$. To denote a scheduling decision, we define a vector $\boldsymbol{b} = \begin{bmatrix} b_{1} \cdots b_{m} \cdots b_{N} \end{bmatrix}$ such that $b_{m}=1$ if $m \in \mathcal{K}$
(digital devices) and $b_{m}=0$ otherwise, while we use a binary variable $c = 1$ to indicate the scheduling of at least one OTA device. Combining the latency of OTA and digital devices, the overall upload latency 
of one FL round is given by
\begin{align}
\tau = c \cdot  \frac{d}{B} + \sum_{m \in \mathcal{N}} \frac{b_m \cdot L_m}{ B \cdot \log_2\left(1 + \text{SNR}_{m}\right)},
\label{Uplink_time}
\end{align}
Let $\hat{\boldsymbol{g}}_{m} \, m \in \mathcal{K}$ denote the estimated local gradient of a digital device at the PS, then the estimate of the global gradient is constructed at the PS as
\begin{align}
    \hat{\boldsymbol{g}} = \frac{1}{N}\left(\sum\limits_{m \in \mathcal{K}} \hat{\boldsymbol{g}}_{m} + \boldsymbol{v}\right)
    \label{grad_recon},
\end{align}
where recall $\boldsymbol{v}  = \frac{\mathbf{y}}{\gamma}$. The estimated global gradient $\hat{\boldsymbol{g}}$ in \eqref{grad_recon} is used in \eqref{GD} in place of ${\boldsymbol{g}}$ to update the FL model. Note that the estimate of the global gradient is unbiased due to zero-mean quantization and white noise in digital and OTA transmissions, respectively. To capture the performance of ADFL, focusing on one round, we define the mean squared error (MSE)  on the global gradient estimated at the PS as
\begin{align}
\text{MSE} \triangleq \mathbb{E} \left[ \left\Vert\boldsymbol{g} -  \hat{\boldsymbol{g}} \right\Vert^2\right],
\label{init_MSE}
\end{align}
where $\boldsymbol{g}$ is the noiseless global gradient, and the expectation here is taken with respect to quantization noise
and white noise at the PS. Recall that the estimated noisy sum of local gradients in OTA aggregation is given as $\boldsymbol{v}  = \sum_{m \in \mathcal{N} \setminus \mathcal{K}}  \boldsymbol{g}_{m} + \mathbf{w}$, where $\mathbf{w} \sim \mathcal{CN}(\boldsymbol{0},\frac{N_0}{\gamma^2}\textbf{I})$, while the reconstructed local gradients of digital devices are subject to quantization noise. Hence, using \eqref{grad_recon}, it is straightforward to compute an upper bound on the MSE as
\begin{align}
 \text{MSE} \leq  d \left(  \sum_{m \in \mathcal{N}} b_m \cdot \left(\frac{\Vert \boldsymbol{g}_{m} \Vert_\infty}{2^{r_{m}} -1}\right)^2+ \, c\cdot \frac{N_0}{{\gamma}^2}\right),
 \label{MSE_simplified}
\end{align}
where the MSE decomposes into respective digital (the first term) and OTA (the second term) MSEs. It can be observed from \eqref{gamma_def} that the device scheduling directly affects $\gamma$, which consequently controls the MSE in \eqref{MSE_simplified}. To capture this, we define a scheduling metric (SM) for each device as
\begin{align}
\text{SM}_{m} = \frac{\vert h_{m}\vert}{\Vert \boldsymbol{g}_{m} \Vert_\infty }. 
\label{SM_def}
\end{align} 
Therefore, for a given scheduling, the PS can compute $\gamma$ with the knowledge of SM$_{m}$  as $\gamma = \sqrt{E_s}\min\limits_{m \in \mathcal{N} \setminus \mathcal{K}} \{\text{SM}_{m}\}$. 
Note that while a device with a low SM$_m$ requires more time to transmit the local gradient over a digital channel, a higher local gradient norm can also help in convergence. Hence, next, we study the MSE minimization to realize less noisy FL updates as a surrogate objective for achieving faster convergence. 
\section{Problem Formulation and Proposed Solution}

It can be seen from \eqref{Uplink_time} that the uplink time in ADFL reduces as more devices are scheduled for OTA transmission; however, such a scheduling approach can potentially reduce $\gamma$ in \eqref{gamma_def}, resulting in worse MSE. On the other hand, the MSE can be improved by scheduling more devices to use the digital scheme at the expense of a higher round delay. 
Owing to this trade-off in each FL round, we now formulate a delay-constrained MSE minimization problem using the upper bound in \eqref{MSE_simplified}:
\begin{gather}
	 \min_{\boldsymbol{b}, \,c\,,\,\{r_m\}}  \; \sum_{m \in \mathcal{N}} b_m \cdot d \left(\frac{\Vert \boldsymbol{g}_{m} \Vert_\infty}{2^{r_m} -1}\right)^2+ c \cdot \frac{ d\, N_0}{{\gamma}^2} \label{Prob_Form} \tag{P1}, \\
	  s.t. \quad \quad \gamma = \sqrt{E_s}\min\limits_{m:b_m = 0} \{\text{SM}_{m}\} \;, \forall m \in \mathcal{N},\\
	       \sum_{m \in  \mathcal{N}} b_m \left(\frac{L_m}{B \log_2(1+\text{SNR}_m)}\right) +  \frac{c \,d}{B} \leq T_\text{max}, \label{T_max_cons} \\
	         r_m \in \{1,2, \cdots\}\;,\,\forall m \in \mathcal{N}: b_m =1 , \label{bit_cons}\\
	        b_m \in \{0,1\} \,,\,\forall m \in \mathcal{N},\ 
	        c=\chi(\sum_{m\in\mathcal N}b_m<N)\label{sched_con}.
\end{gather}
Constraint \eqref{sched_con} in \eqref{Prob_Form} ensures that each device gets scheduled and only to one of the transmission schemes, whereas \eqref{bit_cons} guarantees at least one-bit allocation to each digital device. Finally, \eqref{T_max_cons} requires the round delay to be constrained by $T_\text{max}$, where we assume $T_\text{max} \geq \frac{d}{B}$ for the feasibility of \eqref{Prob_Form}.

It can be verified that \eqref{Prob_Form} is a combinatorial optimization problem, and therefore, not straightforward to solve.
Therefore, we break the problem into two subproblems: 1) scheduling of devices and 2) bits allocation to digital devices.  To address the computational challenge, we first provide a structural property in Theorem \ref{Shed_thm}.
\begin{theorem}
Given any scheduling configuration with $K<N$ digital devices, the $K$ devices with the smallest SM should be scheduled as digital devices for optimal MSE performance.
\label{Shed_thm}
\end{theorem}
A sketch of the proof is provided in the Appendix. Without loss of generality, let devices be sorted in order of non-decreasing $\text{SM}_{m}$. Then, using Theorem \ref{Shed_thm}, we reduce the set $\mathcal{S}$ of $2^N$ possible scheduling configurations to the set $\mathcal{S}^\prime \subset \mathcal{S}$ defined as, $\mathcal{S}^\prime = \{\boldsymbol{b} \mid {b}_m \geq {b}_n \text{ for } m < n, m, n \in \mathcal{N}\}$.
Note that the above theorem provides two key insights. First, for a given scheduling, any device $n$ with SM$_n >$ SM$_{m^\prime}$, where ${m^\prime} = \arg\min\limits_{m \in \mathcal{N} \setminus \mathcal{K}} \{ \text{SM}_m\}$ can be scheduled as an OTA device without degrading the MSE. Further, it also reduces the round delay speeding up the convergence.
Thanks to the SM-based ordering of devices implied by Theorem \ref{Shed_thm}, the optimal scheduling configuration can be obtained using a linear search over $\mathcal{S}^\prime$ with $\mathcal{O}(N + 1)$ complexity.

To solve \eqref{Prob_Form}, we first find a solution to the bit allocation sub-problem for a given scheduling and then optimize over the scheduling configurations using the described approach. 
We approach the bit allocation sub-problem by relaxing the discrete set constraint in \eqref{bit_cons} to a continuous set. 
Thus, letting
$r^\prime_m \geq 0$ be the (continuous) number of bits in excess of 1 allocated to the $m-$th digital device,
we can then approximate the desired (integer) bit allocation as $r_m = \lfloor{r^\prime_m}\rfloor +1$. To find an analytical solution of $r'_m$, we first linearize the denominator of the first term of the objective in \eqref{Prob_Form} around the origin. Since only low SM$_m$ devices (with low channel gain, high gradient norm, or both) use digital channels (Theorem \ref{Shed_thm}), the above small $r_m$ approximation is suitable. Hence, the bit allocation sub-problem can be expressed as
\begin{gather}
	 \min_{\{r^\prime_m\}}  \sum_{m \in \mathcal{N}} b_m \cdot d \left(\frac{\Vert \boldsymbol{g}_{m} \Vert_\infty}{1 + 2\ln(2)\cdot {r^\prime_m}}\right)^2 \label{Prob_Form1} \tag{P1.1}, \\
	 s.t. \quad \quad \eqref{gamma_def}, \\
	       \sum_{m \in  \mathcal{N}} b_m \cdot \left(\frac{64 + d+ d\cdot {r^\prime_m}}{B\cdot \log_2(1+\text{SNR}_m)}\right) + c\cdot \frac{d}{B} \leq T_\text{max}, \label{T_max_cons1}\\	       
	         {r^\prime_m} \geq 0 \;,\,\forall m \in \mathcal{N} \,:
          b_m =1, 
          \label{bit_cons1}
\end{gather}
where we drop the term $\frac{c \,d N_0}{\gamma^2}$ being a constant. It can be shown that, given the scheduling ($b_m$ and $c$), \eqref{Prob_Form1} is a convex optimization problem and, therefore, we refer to duality theory to solve it. The Lagrangian for \eqref{Prob_Form1} is constructed as
\begin{multline}
	 \mathcal{L}({r^\prime_1},\cdots, {r^\prime_n},\lambda) 	 = \sum_{m \in \mathcal{N}} b_m \cdot d \left(\frac{\Vert \boldsymbol{g}_{m} \Vert_\infty}{1 + 2\ln(2)\cdot {r^\prime_m}}\right)^2 + \\  \lambda\left(\sum_{m \in \mathcal{N}} b_m \cdot \frac{64 + d+ d\cdot {r^\prime_m}}{B\cdot \log_2(1+\text{SNR}_m)} +\frac{ c\;d}{B}-T_\text{max}\right). \label{Lagrangian}
\end{multline}

By optimizing over $r^\prime_m\geq 0$, we obtain
\begin{align}
	{r^\prime_m}^*(\lambda) =\max \{A_m(\lambda), 0\}\label{r_star},
\end{align} 
where we have defined 
$$A_m(\lambda)\triangleq \frac{1}{2 \ln(2)} \left(\sqrt[3]{\frac{\lambda_m}  \lambda}  -1 \right)$$
and
\begin{equation} \lambda_m \triangleq 4\cdot \ln(2)\cdot \left\Vert \boldsymbol{g}_{m}\right\Vert^2_\infty \cdot B \log_2(1 + \text{SNR}_m).
\label{lambda_m}
\end{equation}

The dual Lagrangian function $q(\lambda)$ is expressed by replacing the expression of 
${r^\prime_m}^*(\lambda)$ into \eqref{Lagrangian}.
Next, we maximize the dual function to obtain $\lambda^*$, i.e., $\lambda^*=\arg\max_{\lambda\geq0} q(\lambda)$. However, due to the non-differentiability of the  $\max \{A_m(\lambda), 0\}$ term in \eqref{r_star}, a closed-form solution of $\lambda^*$ cannot be directly obtained.
To address this challenge, we partition the interval $\lambda\geq 0$ into regions corresponding to activations of ${r^\prime_m}^*$ in \eqref{r_star}.
To this end, note that 
${r_m^\prime}^* = A_m(\lambda) > 0$ if and only if $\lambda < \lambda_m$.
Without loss of generality,
we reorder the $K$ digital devices such that $\lambda_1 \leq \cdots \leq \lambda_K$, then it follows that $r_n>0,\forall n>m$ and $r_n=0,\forall n\leq m$ for $ \lambda_m \leq \lambda < \lambda_n$.
Thus, we compute $\lambda^*$
by optimizing it over a differentiable interval $(\lambda_m,\lambda_{m+1})$,
instead of directly searching over the entire interval $\lambda>0$. Specifically, for a given scheduling to be feasible, first the round delay constraint in \eqref{T_max_cons1} is verified by setting $\forall m \in \mathcal{K},\,{r^\prime_m} = 0$ (corresponding to $\lambda \geq \max\limits_{m \in \mathcal{K}} \lambda_m$). 
Note that if the given scheduling configuration is infeasible, we discard the configuration and proceed with another configuration that does not violate \eqref{T_max_cons1}.
For a feasible scheduling configuration, letting $q^\prime(x^{\pm})\triangleq \frac{d(q(\lambda))}{ d \lambda} \rvert_{\lambda =x^\pm}$
denote the left ($x^-$) and right ($x^+$) derivatives at $x$,
either $\exists\, m \in \{1,\cdots,K\}$ such that $q^\prime(\lambda_{m-1}^+) > 0 $ and $q^\prime(\lambda_m^-) < 0$: in this case, $\lambda^*$
can be obtained as the unique solution of
$q^\prime(\lambda^*)=0$
in $(\lambda_{m-1},\lambda_{m})$, 
due to the concavity of $q(\lambda)$; otherwise $\exists \,m$ such that $q^\prime(\lambda_m^-) > 0$ and $q^\prime(\lambda_m^+) < 0$, for which $\lambda^* = \lambda_{m}$. With the value of $\lambda^*$ thus obtained, we then find the optimal bit allocation via \eqref{r_star}.
Substituting the optimized $r^*_m$ in \eqref{Lagrangian} gives a bound on the MSE with the optimal bit allocation.
Finally, a linear search is performed over the reduced set of configurations $\mathcal{S}^\prime$ to minimize the obtained MSE, yielding the optimal scheduling configuration $\boldsymbol{b}^*$.

\section{Numerical Results}

We perform numerical experimentation to evaluate the performance of ADFL. Specifically, a classification task is performed in a FL setting on the widely used MNIST dataset \cite{MNIST}, which consists of C = 10 classes ranging from digit “0” to “9” to classify 28 x 28 pixels images of handwritten digits on a single-layer neural network. The FL is performed over a network of $N= 10$ devices uniformly deployed
within a radius of $r_\text{max} = 200 $ m from the PS.
The devices communicate with a bandwidth $B$ = 1 MHz over a carrier frequency $f_c =$ 2.4 GHz with fixed transmission power $P_\text{tx} =$ 20 dBm. The noise power spectral density at the receivers is $N_0 = -174$dBmW/Hz. The average path loss $\Lambda_m$ between devices and the PS follows the log-distance path loss model with path loss exponent $\beta = 2.2$, and the reference distance is assumed to be 1 meter.  
The optimization parameter $\boldsymbol{\theta
} \in \mathbb{R}^{7850}$ is given as, $\boldsymbol{\theta}^T =$  
$\begin{bmatrix}
{\boldsymbol{\theta}^{(0)}}^T,\cdots,{\boldsymbol{\theta}^{(9)}}^T \end{bmatrix}$. The classification task is performed using regularized cross-entropy local loss functions at each device, given by
\begin{align*}
    \phi((\boldsymbol{x},\ell );\boldsymbol{\theta})= \frac{1}{2}\Vert \boldsymbol{\theta}\Vert^2 -\ln \left(\frac{\exp{\{ \boldsymbol{x}^T\boldsymbol{\theta}^{(l)}\}}}{\sum_{c = 0}^9{\exp{\{ \boldsymbol{x}^T\boldsymbol{\theta}^{(c)}\}}}}\right).
\end{align*}
\begin{figure*}[t]
\centering
\subfloat[\centering i.i.d. data points distribution, $N =10$ devices, $T_\text{max} \approx 63$ ms.]{{\includegraphics[width=0.457\textwidth]{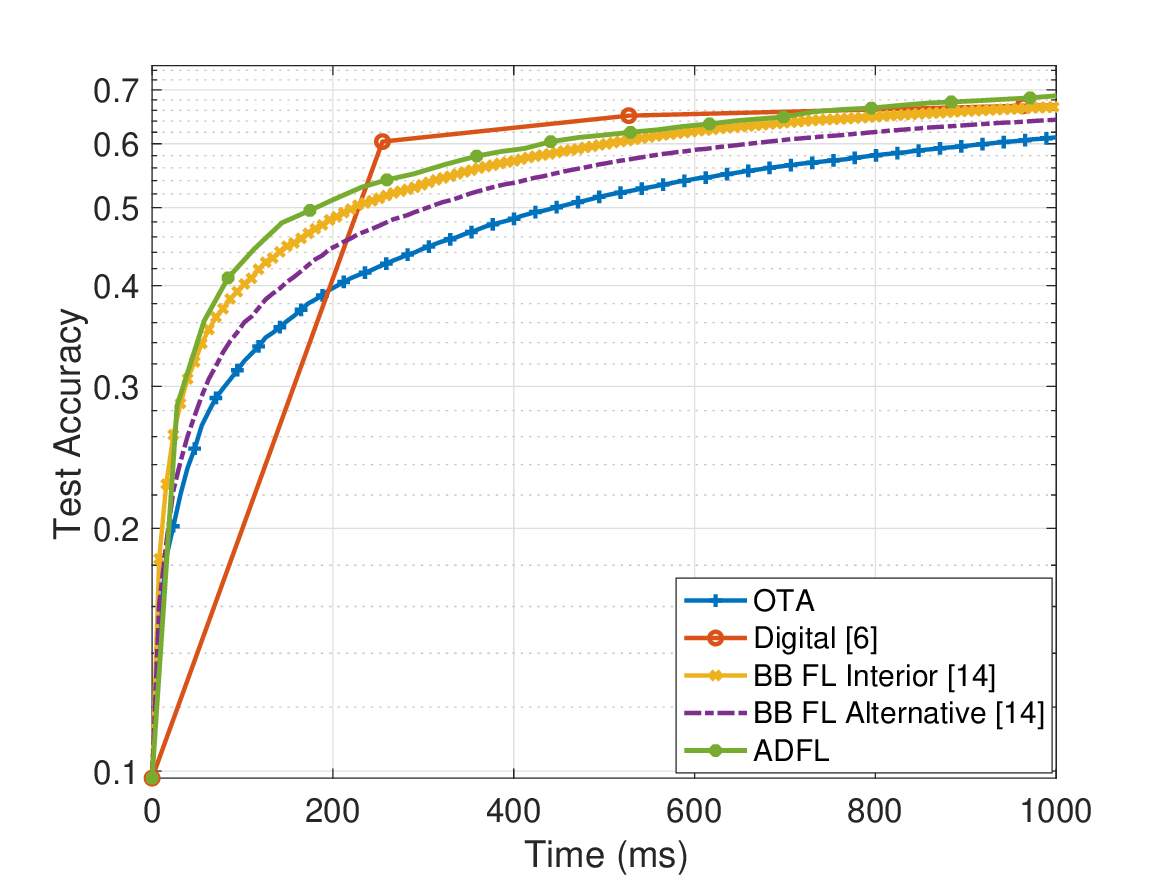} }}%
\qquad
\subfloat[\centering non-i.i.d. data points distribution, $N =10$ devices, $T_\text{max} \approx 20$ ms.]{{\includegraphics[width=0.47\textwidth]{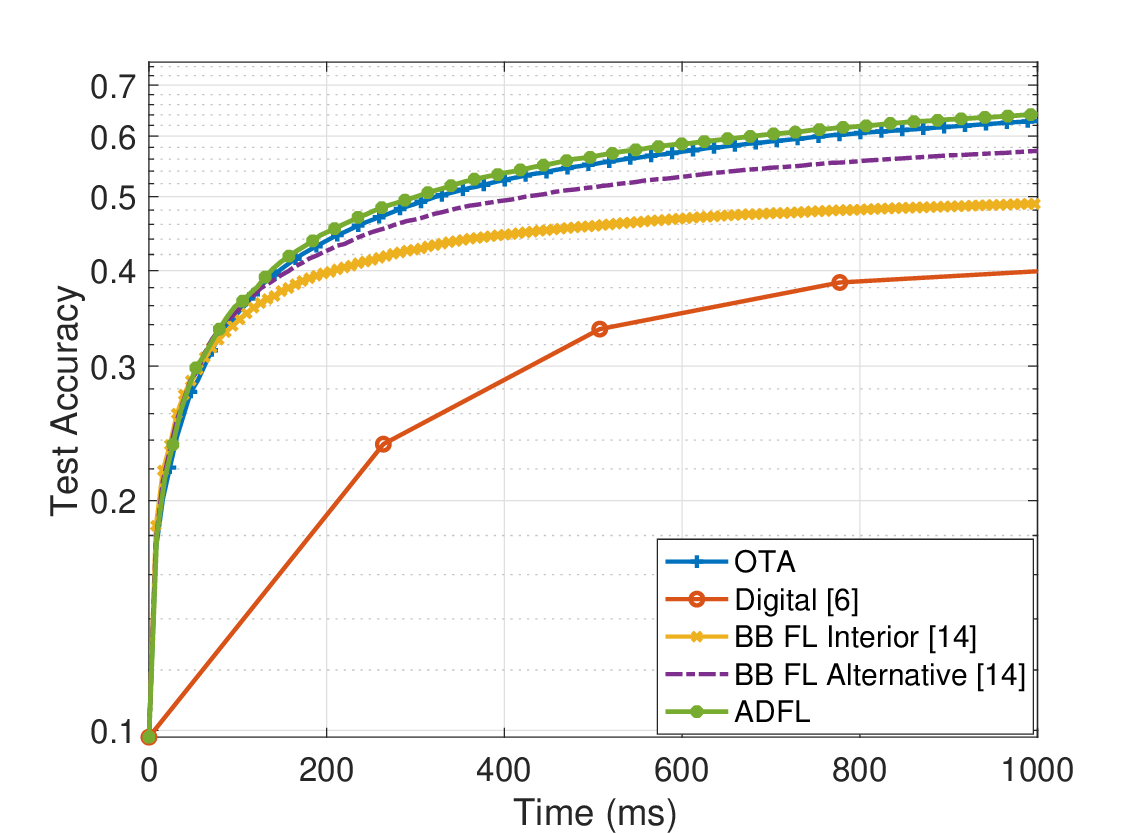} }}%
\caption{Test accuracy comparison of ADFL with existing OTA and digital schemes for i.i.d. and non-i.i.d. settings.}%
\label{FL_results}
\end{figure*}
We perform experiments for both i.i.d. and non-i.i.d. training data distribution scenarios with $\sum\limits_{m \in \mathcal{N}} \vert \mathcal{D}_m\vert = 1000$ training data points. For the i.i.d. case, each device has 100 samples chosen uniformly at random from the overall training data without replacement. To create a non-i.i.d. setting, 
we allow each device to access only the data points of two class labels chosen at random, such that no more than two devices have access to the data points of a particular class. A mini-batch of size $\vert\mathcal{B}_m\vert =50$ samples is used by each device to compute the local gradients.

To demonstrate the effectiveness of ADFL, we also compare it with the following existing schemes: 1) OTA scheme \cite{OTA_FL}, in which each device is scheduled to use OTA aggregation to upload the gradients;\footnote{While \cite{OTA_FL} randomly samples a fraction of total devices to participate, we assume here the participation of each device in each round for fair comparison.} 2) Digital scheme, used in \cite{Sched_policies}, which schedules a device based on the norm of the local gradient and uploading time; 3) BB FL Interior, proposed in \cite{BB_FL}, which schedules only the devices that are within a radius $R_\text{in} < r_\text{max}$ from the PS; and 4) BB FL Alternative, proposed in \cite{BB_FL}, which alternates between scheduling every device and BB FL Interior. Specifically, Digital scheme \cite{Sched_policies} schedules device $m$ with probability $p_m \propto \frac{\sqrt{\rho}\Vert \boldsymbol{g}_m\Vert}{(1-\rho)T_m}$, where $\rho \in [0,1]$ is the weight to schedule based on the norm and the upload time. The local gradient is quantized using $r_m$ bits and sent over a digital channel utilizing $T_m = \frac{d r_m}{B \log_2\left(1 + \text{SNR}_{m}\right)}$ time. To obtain the best performance of the schemes, we set $K=1$, $r_m = 16$, and $\rho = 5 \times 10^{-4}$
for Digital scheme, whereas, we set $R_{\text{in}} = 0.6\,r_{\text{max}}$ for BB FL Interior and BB FL Alternative schemes, as demonstrated in \cite{Sched_policies} and \cite{BB_FL}
. Finally, while $T_\text{max}$ can be treated as a tunable hyperparameter in each ADFL round, we keep it fixed throughout the FL running time.  

In Fig. \ref{FL_results}, we show the test accuracy attained for a FL running time of $1000$ ms, averaged over channel and deployment realizations. Fig. \ref{FL_results}a plots the i.i.d. case with $T_{\text{max}} = 8\,\tau_\text{OTA} \approx 63 $ ms, where recall $\tau_\text{OTA} = \frac{d}{B}$. It can be seen that, although the digital scheme
gives the best performance initially, 
thanks to it being less noisy, the accuracy of ADFL grows faster over time, resulting in better final performance. Furthermore, note that while each update of the digital scheme takes more time, each device having \textit{similar} data reduces the need for collaboration, accounting for the accuracy achieved for $K= 1$. In addition, we observe that ADFL provides higher accuracy than OTA, BB FL Interior, and BB FL Alternative throughout the running time. Among the OTA schemes, BB FL Interior shows the best performance due to a similar reason as given for the digital scheme. Finally, we notice that ADFL achieves the same accuracy as BB FL Alternative and OTA  in 38\% and 50\% less time, respectively.

In the non-i.i.d. setting depicted in Fig. \ref{FL_results}b, since every device brings relatively \textit{unique} information, scheduling more devices gives better performance, in general. We observe that ADFL outperforms all other schemes during the entire time range. A choice of $T_\text{max} = 2.5 \,\tau_{\text{OTA}} \approx 20 $ ms yields an improvement over the OTA scheme, thanks to digital channels letting \textit{weak} devices participate without deteriorating the OTA transmission variance. It is worth mentioning that in the non-i.i.d. setting, the gain achieved by ADFL over OTA is less than in the i.i.d. case. This difference arises because scheduling a subset of devices to transmit over digital channels conveys less information and consumes more time compared to the i.i.d. scenario. For the digital and BB FL Interior schemes, since some devices might not be scheduled, the model is unable to accurately predict samples of unseen classes, explaining the performance degradation. Note that ADFL requires $46 \%$ less time than BB FL Alternating, the best among other schemes, for the same accuracy. Overall, it can be observed that ADFL, by leveraging digital channels for the \textit{weak} (low SM) devices, consistently performs well in both i.i.d. and non-i.i.d. settings. 

\section{Conclusion}

In this paper, we presented ADFL, a scheme that takes advantage of both OTA computation and orthogonal digital channels to realize FL over a wireless network. To prove the effectiveness of ADFL in balancing the trade-off between MSE and FL round delay, we formulated a per-round device scheduling and digital device bit-allocation problem. To solve the problem, we provided a method to obtain a closed-form solution for the approximated bit allocation problem. Moreover, we have also shown that the combinatorial device scheduling problem can be optimally solved with a linear search. To support our analysis, we further performed numerical evaluations, demonstrating that, while existing schemes perform well in specific settings, ADFL consistently exhibits superior performance across multiple settings.

\appendix
\textit{Proof sketch of Theorem \ref{Shed_thm}.} Without loss of generality, we assume that the $N$ devices are sorted according to SM so that $\text{SM}_1 \leq \cdots \leq \text{SM}_K \leq  \cdots \leq \text{SM}_N$.  Then, for a given scheduling configuration with $N-K$ OTA devices, let $m^\prime$ denote the index of the OTA device ($b_{m^\prime} = 0$) with the smallest SM, i.e., $\text{SM}_{m^\prime} \leq \text{SM}_m, \forall m \in \mathcal{N}, \text{such that } b_m =0$. It follows from \eqref{gamma_def} that $\gamma = \sqrt{E_s}\,\text{SF}_{m^\prime}$. Hence, the MSE in \eqref{MSE_simplified} is given by $d \left(  \sum\limits_{m \in \mathcal{N}} b_m \left(\frac{\Vert \boldsymbol{g}_{m} \Vert_\infty}{2^{r_{m}} -1}\right)^2+ \, \frac{N_0}{E_s \, \text{SF}^2_{m^\prime}}\right)$. Clearly, the second term shows that the MSE corresponding to OTA transmission is only attributed to device $m^\prime$. Consequently, if $\exists\, n \in \mathcal{N}$ such that $\text{SF}_{m^\prime} \leq \text{SF}_n \,, b_n =1$, then it can be seen that scheduling $n$ as an OTA device ($b_n = 0$) decreases the first term without increasing the second term of the MSE, resulting in a configuration with better MSE and reduced round latency. Therefore, the above argument can be generalized to conclude that scheduling any device having index $K+1$ to $N$ as a digital device results in a strictly suboptimal device scheduling configuration.

\bibliographystyle{IEEEtran} 
\bibliography{Refs} 

\end{document}